%% file: WAFR.tex
\RequirePackage{amsmath}    
\documentclass{llncs}

\usepackage{times}
\usepackage{amsmath}    
\usepackage{amsfonts}
\usepackage{amssymb}
\usepackage{graphicx}
\usepackage{color}
\usepackage{booktabs}
\usepackage[normalem]{ulem} 

\usepackage{todonotes}

\usepackage{ifthen}
\newboolean{include-notes}
\setboolean{include-notes}{true}

\newcommand{\adnote}[1]{\ifthenelse{\boolean{include-notes}}%
 {\textcolor{blue}{\textbf{AD: #1}}}{}}
 \newcommand{\jfnote}[1]{\ifthenelse{\boolean{include-notes}}%
 {\textcolor{green}{\textbf{JF: #1}}}{}}
 \newcommand{\jbhnote}[1]{\ifthenelse{\boolean{include-notes}}%
 {\textcolor{purple}{\textbf{JBH: #1}}}{}}
  \newcommand{\clnote}[1]{\ifthenelse{\boolean{include-notes}}%
 {\textcolor{orange}{\textbf{CL: #1}}}{}}
 
 \newcommand{\add}{\textcolor[rgb]{0,0.0,0.0}}

 \newcommand{\AC}{predictable}
 \newcommand{\ACty}{predictability}
 \newcommand{\ACzero}{optimal}
 \newcommand{\ACone}{1-predictable}
 \newcommand{\ACtwo}{2-predictable}
 \newcommand{\bg}{\mathbf{a}}

\input{h_bib}
\input{h_com}

\usepackage{hyperref}
\hypersetup{bookmarksopen,bookmarksnumbered,
pdfpagemode=UseOutlines,
colorlinks=true,
linkcolor=blue,
anchorcolor=blue,
citecolor=blue,
filecolor=blue,
menucolor=blue,
urlcolor=blue,
}

\renewcommand{\subsubsection}[1]{\vspace{0.1em}\noindent\textbf{#1}}

\title{Generating Plans that Predict Themselves}
\author{{Jaime F.~Fisac\inst{1}\thanks{These authors contributed equally.},
    Chang Liu\inst{2}$^{\star}$,
    Jessica B.~Hamrick\inst{3}$^{\star}$,}\\
    {Shankar Sastry\inst{1},
    J. Karl Hedrick\inst{2},
    Thomas L.~Griffiths\inst{3},
    Anca D.~Dragan\inst{1}}
}
\institute{Department of Electrical Engineering and Computer Sciences\and Department of Mechanical Engineering \and Department of Psychology\\ University of California, Berkeley,
CA 94720 U.S.A.
\email{\{jfisac,changliu,jhamrick,shankar\_sastry,tom\_griffiths,anca\}\\@berkeley.edu
}
}

\begin{document}
\maketitle
\pagestyle{plain}

\input{results/mturk_theoretical_predictability}
\input{results/mturk_accuracy}
\input{results/mturk_preferences_over_time}
\input{results/mturk_final_rankings}
\input{results/user_study_completion_rate}
\input{results/user_study_task_sequence}
\input{results/user_study_preferences}
\input{results/user_study_survey_responses}

\input{abstract}

\input{introduction}

\input{formulation}
	
\input{online-experiment}

\input{user-study}

\input{discussion}
\printbibliography{}

\end{document}

%% file: h_bib.tex
\usepackage{url}
\usepackage{hyperref}
\usepackage[%
    style=numeric-comp,
    sorting=nyt,
    backend=biber,
    sortcites=true,
    doi=false,
    firstinits=true,
    hyperref,
    isbn=false,
    eprint=false,
    minbibnames=1,   
    maxbibnames=3,
    block=none]
    {biblatex}
    
    \renewbibmacro{in:}{}
    \AtEveryBibitem{%
      \clearlist{language}%
      \clearfield{pages}
      \iffieldequalstr{entrykey}{Liu2016}
        {\setcounter{minnames}{3}} 
        {\setcounter{minnames}{1}} 
      }

\DeclareBibliographyCategory{needsurl}
\newcommand{\entryneedsurl}[1]{\addtocategory{needsurl}{#1}}
\renewbibmacro*{url+urldate}{%
  \ifcategory{needsurl}{
    \printfield{url}%
    \iffieldundef{urlyear}
      {}
      {\setunit*{\addspace}%
       \printurldate}}
    {}}

\bibliography{library,library2}
\entryneedsurl{Mitchell2004}

%% file: h_com.tex
\newcommand{\A}{\mathcal{A}}

%% file: results/mturk_theoretical_predictability.tex
\newcommand{\PredictabilityVsAccuracy}[0]{$r=0.87,\ 95\%\ \mathrm{CI}\ [0.81, 0.91]$}
\newcommand{\PredictabilityVsAccuracyShort}[0]{$r=0.87$}

%% file: results/mturk_accuracy.tex
\newcommand{\AccuracyConditionEffect}[0]{$F(2, 10299)=1894.75,\ p < 0.001$}
\newcommand{\AccuracyRobotEffect}[0]{$F(2, 42)=6.59,\ p < 0.01$}
\newcommand{\AccuracyInteraction}[0]{$F(4, 10299)=554.00,\ p < 0.001$}
\newcommand{\AccuracyConditionZeroLegibleOnevsLegibleTwo}[0]{$t(49)=0.03,\ p = 1.00$}
\newcommand{\AccuracyConditionZeroLegibleOnevsOptimal}[0]{$t(49)=7.15,\ p < 0.001$}
\newcommand{\AccuracyConditionZeroLegibleTwovsOptimal}[0]{$t(49)=7.12,\ p < 0.001$}
\newcommand{\AccuracyConditionOneLegibleOnevsLegibleTwo}[0]{$t(49)=-12.29,\ p < 0.001$}
\newcommand{\AccuracyConditionOneLegibleOnevsOptimal}[0]{$t(49)=-4.72,\ p < 0.001$}
\newcommand{\AccuracyConditionOneLegibleTwovsOptimal}[0]{$t(49)=7.57,\ p < 0.001$}
\newcommand{\AccuracyConditionTwoLegibleOnevsLegibleTwo}[0]{$t(50)=1.85,\ p = 0.56$}
\newcommand{\AccuracyConditionTwoLegibleOnevsOptimal}[0]{$t(50)=-6.67,\ p < 0.001$}
\newcommand{\AccuracyConditionTwoLegibleTwovsOptimal}[0]{$t(50)=-8.52,\ p < 0.001$}
\newcommand{\AccuracyLegibleOneCondZerovsCondOne}[0]{$t(10299)=39.21,\ p < 0.001$}
\newcommand{\AccuracyLegibleOneCondZerovsCondTwo}[0]{$t(10299)=47.00,\ p < 0.001$}
\newcommand{\AccuracyLegibleOneCondOnevsCondTwo}[0]{$t(10299)=8.31,\ p < 0.001$}
\newcommand{\AccuracyLegibleTwoCondZerovsCondOne}[0]{$t(10299)=3.26,\ p < 0.05$}
\newcommand{\AccuracyLegibleTwoCondZerovsCondTwo}[0]{$t(10299)=52.28,\ p < 0.001$}
\newcommand{\AccuracyLegibleTwoCondOnevsCondTwo}[0]{$t(10299)=49.06,\ p < 0.001$}
\newcommand{\AccuracyOptimalCondZerovsCondOne}[0]{$t(10299)=4.57,\ p < 0.001$}
\newcommand{\AccuracyOptimalCondZerovsCondTwo}[0]{$t(10299)=7.16,\ p < 0.001$}
\newcommand{\AccuracyOptimalCondOnevsCondTwo}[0]{$t(10299)=2.65,\ p = 0.11$}

%% file: results/mturk_preferences_over_time.tex
\newcommand{\PreferenceRobot}[0]{$\chi^2(2)=13.66,p < 0.01$}
\newcommand{\PreferenceCondition}[0]{$\chi^2(2)=4.67,p = 0.10$}
\newcommand{\PreferenceTrial}[0]{$\chi^2(1)=9.30,p < 0.01$}
\newcommand{\PreferenceRobotCondition}[0]{$\chi^2(4)=20.26,p < 0.001$}
\newcommand{\PreferenceRobotTrial}[0]{$\chi^2(2)=24.68,p < 0.001$}
\newcommand{\PreferenceConditionTrial}[0]{$\chi^2(2)=16.07,p < 0.001$}
\newcommand{\PreferenceRobotConditionTrial}[0]{$\chi^2(4)=39.43,p < 0.001$}
\newcommand{\PrefsConditionZeroLegibleOnevsLegibleTwo}[0]{$z=3.86,\ p < 0.01$}
\newcommand{\PrefsConditionZeroLegibleOnevsOptimal}[0]{$z=-13.22,\ p < 0.001$}
\newcommand{\PrefsConditionZeroLegibleTwovsOptimal}[0]{$z=-14.56,\ p < 0.001$}
\newcommand{\PrefsConditionOneLegibleOnevsLegibleTwo}[0]{$z=14.00,\ p < 0.001$}
\newcommand{\PrefsConditionOneLegibleOnevsOptimal}[0]{$z=12.97,\ p < 0.001$}
\newcommand{\PrefsConditionOneLegibleTwovsOptimal}[0]{$z=-4.22,\ p < 0.001$}
\newcommand{\PrefsConditionTwoLegibleOnevsLegibleTwo}[0]{$z=2.26,\ p = 0.29$}
\newcommand{\PrefsConditionTwoLegibleOnevsOptimal}[0]{$z=7.44,\ p < 0.001$}
\newcommand{\PrefsConditionTwoLegibleTwovsOptimal}[0]{$z=5.40,\ p < 0.001$}
\newcommand{\PrefsLegibleOneCondZerovsCondOne}[0]{$z=-13.09,\ p < 0.001$}
\newcommand{\PrefsLegibleOneCondZerovsCondTwo}[0]{$z=-7.26,\ p < 0.001$}
\newcommand{\PrefsLegibleOneCondOnevsCondTwo}[0]{$z=6.97,\ p < 0.001$}
\newcommand{\PrefsLegibleTwoCondZerovsCondOne}[0]{$z=0.20,\ p = 1.00$}
\newcommand{\PrefsLegibleTwoCondZerovsCondTwo}[0]{$z=-8.19,\ p < 0.001$}
\newcommand{\PrefsLegibleTwoCondOnevsCondTwo}[0]{$z=-8.03,\ p < 0.001$}
\newcommand{\PrefsOptimalCondZerovsCondOne}[0]{$z=13.11,\ p < 0.001$}
\newcommand{\PrefsOptimalCondZerovsCondTwo}[0]{$z=13.31,\ p < 0.001$}
\newcommand{\PrefsOptimalCondOnevsCondTwo}[0]{$z=0.60,\ p = 1.00$}

%% file: results/mturk_final_rankings.tex
\newcommand{\RankingsRobot}[0]{$\chi^2(2)=41.38,p < 0.001$}
\newcommand{\RankingsCondition}[0]{$\chi^2(2)=12.97,p < 0.01$}
\newcommand{\RankingsRobotCondition}[0]{$\chi^2(4)=88.52,p < 0.001$}
\newcommand{\RankingsConditionZeroLegibleOnevsLegibleTwo}[0]{$z=-4.40,\ p < 0.001$}
\newcommand{\RankingsConditionZeroLegibleOnevsOptimal}[0]{$z=3.46,\ p < 0.01$}
\newcommand{\RankingsConditionZeroLegibleTwovsOptimal}[0]{$z=5.60,\ p < 0.001$}
\newcommand{\RankingsConditionOneLegibleOnevsLegibleTwo}[0]{$z=-6.54,\ p < 0.001$}
\newcommand{\RankingsConditionOneLegibleOnevsOptimal}[0]{$z=-2.18,\ p = 0.27$}
\newcommand{\RankingsConditionOneLegibleTwovsOptimal}[0]{$z=6.62,\ p < 0.001$}
\newcommand{\RankingsConditionTwoLegibleOnevsLegibleTwo}[0]{$z=-1.33,\ p = 0.84$}
\newcommand{\RankingsConditionTwoLegibleOnevsOptimal}[0]{$z=-4.85,\ p < 0.001$}
\newcommand{\RankingsConditionTwoLegibleTwovsOptimal}[0]{$z=-3.85,\ p < 0.01$}

%% file: results/user_study_completion_rate.tex
\newcommand{\CompletionRobot}[0]{$\chi^2(1)=11.17,p < 0.001$}
\newcommand{\CompletionLegibleOnevsOptimal}[0]{$z=3.34,\ p < 0.001$}

%% file: results/user_study_task_sequence.tex
\newcommand{\UserStudyAccuracyRobot}[0]{$\chi^2(1)=9.49,p < 0.01$}
\newcommand{\UserStudyAccuracyLegibleOnevsOptimal}[0]{$z=3.08,\ p < 0.01$}

%% file: results/user_study_preferences.tex
\newcommand{\UserStudyPreferencesRobot}[0]{$\chi^2(1)=7.14,p < 0.01$}
\newcommand{\UserStudyDifficultyRobot}[0]{$F(2, 39)=14.63,\ p < 0.001$}
\newcommand{\UserStudyDifficultyLegibleOnevsOptimal}[0]{$t(26)=-3.54,\ p < 0.01$}
\newcommand{\UserStudyDifficultyLegibleOnevsSelf}[0]{$t(26)=-5.31,\ p < 0.001$}
\newcommand{\UserStudyDifficultyOptimalvsSelf}[0]{$t(26)=-1.77,\ p = 0.20$}

%% file: results/user_study_survey_responses.tex
\newcommand{\UserStudySurveyRobot}[0]{$F(1, 117)=16.42,\ p < 0.001$}
\newcommand{\UserStudySurveyMeasure}[0]{$F(4, 117)=5.45,\ p < 0.001$}
\newcommand{\UserStudySurveyRobotMeasure}[0]{$F(4, 117)=0.63,\ p = 0.64$}
\newcommand{\UserStudySurveyLegibleOnevsOptimal}[0]{$t(117)=4.05,\ p < 0.001$}
\newcommand{\UserStudySurveyCapabilityvsFluency}[0]{$t(117)=3.06,\ p < 0.05$}
\newcommand{\UserStudySurveyCapabilityvsLegibility}[0]{$t(117)=3.24,\ p < 0.05$}
\newcommand{\UserStudySurveyCapabilityvsPredictability}[0]{$t(117)=0.67,\ p = 0.98$}
\newcommand{\UserStudySurveyCapabilityvsTrust}[0]{$t(117)=3.59,\ p < 0.01$}
\newcommand{\UserStudySurveyFluencyvsLegibility}[0]{$t(117)=0.18,\ p = 1.00$}
\newcommand{\UserStudySurveyFluencyvsPredictability}[0]{$t(117)=-2.39,\ p = 0.14$}
\newcommand{\UserStudySurveyFluencyvsTrust}[0]{$t(117)=0.53,\ p = 0.99$}
\newcommand{\UserStudySurveyLegibilityvsPredictability}[0]{$t(117)=-2.57,\ p = 0.09$}
\newcommand{\UserStudySurveyLegibilityvsTrust}[0]{$t(117)=0.34,\ p = 1.00$}
\newcommand{\UserStudySurveyPredictabilityvsTrust}[0]{$t(117)=2.92,\ p < 0.05$}

%% file: abstract.tex
\begin{abstract}
Collaboration requires coordination, and we coordinate by anticipating our teammates' future actions and adapting to their plan. In some cases, our teammates' actions early on can give us a clear idea of what the remainder of their plan is, i.e. what action sequence we should expect. In others, they might leave us less confident, or even lead us to the wrong conclusion. Our goal is for robot actions to fall in the first category: we want to enable robots to select their actions in such a way that human collaborators can easily use them to correctly anticipate what will follow. While previous work has focused on finding initial plans that convey a set goal, here we focus on finding two portions of a plan such that the initial portion conveys the final one. We introduce $t$-\ACty{}: a measure that quantifies the accuracy and confidence with which human observers can predict the remaining robot plan from the overall task goal and the observed initial $t$ actions in the plan. We contribute a method for generating $t$-predictable plans: we search for a full plan that accomplishes the task, but in which the first $t$ actions make it as easy as possible to infer the remaining ones. The result is often different from the most efficient plan, in which the initial actions might leave a lot of ambiguity as to how the task will be completed. Through an online experiment and an in-person user study with physical robots, we find that our approach outperforms a traditional efficiency-based planner in objective and subjective collaboration metrics. 
\end{abstract}

%% file: introduction.tex

\begin{figure}[t!]
\begin{center}
\includegraphics[width=\columnwidth,trim=0mm 6mm 0mm 6mm, clip=true] {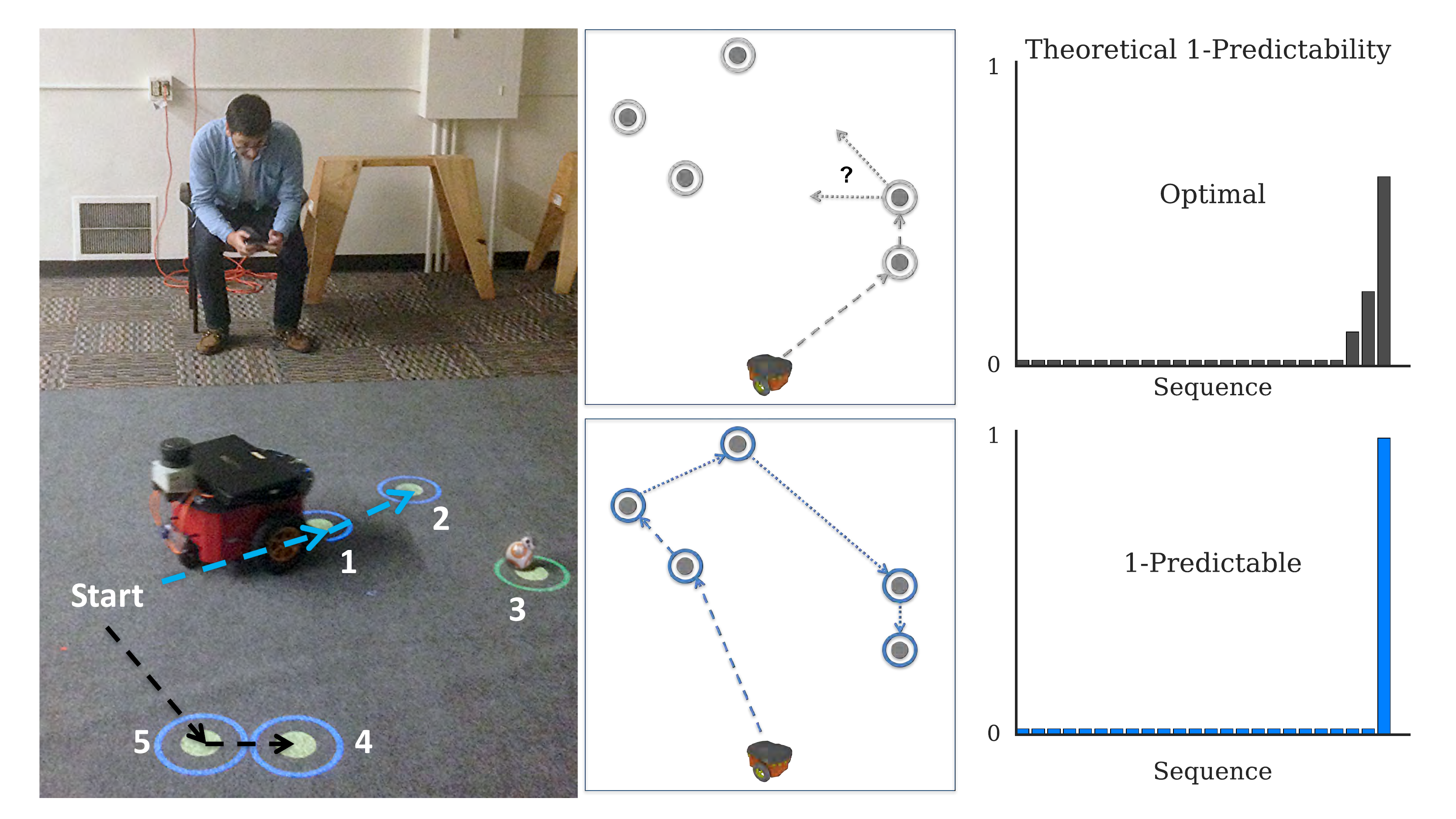}
\caption{Left: A participant controls the toy robot to clear the target that he believes will be third in the autonomous robot's sequence. Center: Schematic of the trajectory followed by the optimal (top) and 1-\AC{} (bottom) planner. Right: Theoretical $1$-\ACty{} of candidate sequences after the robot has visited a single target.
}
\label{fig:front}
\end{center}
\end{figure}

\section{Introduction}\label{introduction}

With robots stepping out of industrial cages and into mixed workspaces shared with human beings, human-robot collaboration is becoming more and more important \cite{Fong2005,Dias2008,Shah2010}. 
There is unfortunately a history of serious and sometimes tragic failures in human-automation systems due to inadequate interaction between machines and their operators \cite{McRuer1995,Sarter1997,Saffarian2012a}. The most common reasons for this are mode confusion and ``automation surprises'', i.e. \emph{misalignments} between what the automated agent is planning to do and what the human believes it is planning to do.

Our goal is to eliminate such misalignments: we want humans to be able to infer what a robot is planning to do during a collaborative task.
This is important even beyond safety reasons \cite{Alami2006}, because it enables the human to adapt their own actions to the robot and more effectively achieve common goals \cite{tomasello2005understanding,vesper2010minimal,pezzulo2013human}.

Traditionally, human-robot collaboration work has focused on inferring human plans and adapting the robot's plan in response \cite{Nikolaidis2013,Liu2016}. 
In contrast, here we are interested in the \emph{opposite}---making sure that humans can make these inferences about the robot.
We envision that ultimately these two approaches will need to work in conjunction to achieve fluent collaboration.
Though it is possible for robots to explicitly communicate their plans via language or text, we focus here on what the beginning of the plan itself implies because 1) people make inferences based on actions \cite{Baker2009}, and 2) certain scenarios such as the outdoors or busy factories might render explicit channels undesirable. 

We introduce a property of a robot plan that we call \emph{$t$-\ACty{}}: a plan is $t$-\AC{} if a human can infer the robot's remaining actions in a task from having observed only the first $t$ actions and from knowing the robot's overall goal in the task. 
We make the following contributions based on this property:

\subsubsection{An algorithm for generating $t$-\AC{} plans.} To generate robot plans that are $t$-\AC{}, we introduce a model for how humans might infer future plans, building on models for action interpretation and plan recognition \cite{Charniak1993,Baker2009}. We then propose a planning algorithm that optimizes for $t$-\ACty{}, i.e. it optimizes plans for how easy it will be for a human to infer the last actions from the initial ones.

Prior work has focused on legibility: generating trajectories that communicate a \emph{set goal} via the initial robot \emph{motion} \cite{Takayama2011,Dragan2014,Szafir2014,Gielniak2011}. This is less useful in task planning situations, where the human already knows the task goal. Instead, $t$-\ACty{} is about communicating the \emph{sequence of future actions} that the robot will take given a \emph{known goal}. The important difference is that these actions are not set \emph{a priori}: optimizing for $t$-\ACty{} means \emph{changing not just the initial, but also the final part} of a plan. It is about finding a final part that is easy to communicate, along with an initial part that communicates it.

Our insight is that initial actions can be used to clarify what future actions will be. We find that in many situations, the robot can select initial actions that might seem somewhat surprising at first, but that make the remaining sequence of actions trivial to anticipate (or ``auto-complete''). 
Fig.~\ref{fig:front} shows an example. If the robot acts optimally for the shortest path through all targets, it will go to target 5 first, making target 4 an obvious candidate as a next step, but leaving the future ordering of targets 1, 2, 3 somewhat ambiguous (high probability on  multiple sequences). On the other hand, if the robot instead chooses target 1, users can with high probability predict that the remaining sequence will be 2-3-4-5.

\subsubsection{An online user study testing that we can generate $t$-\AC{} plans.} We conduct an online user study in which participants observe a robot's partial plan, and anticipate the order of the remaining actions. We find that participants are significantly better at anticipating the correct order when the robot is planning for $t$-\ACty{}. We also find a \PredictabilityVsAccuracyShort{} correlation between our model's prediction of the probability of success, and the participants' actual success rate.

\subsubsection{An in-person user study testing the implications on human-robot collaboration.} Armed with evidence that we can make plans $t$-\AC{}, we move on to an in-person study that puts participants in a collaborative task with the robot, and study the advantages of $t$-\ACty{} on objective and subjective collaboration metrics. We find that participants were more effective at the task, and prefer to work with a $t$-\AC{} robot than with an optimal robot.

%% file: formulation.tex
\section{Defining and Optimizing for $t$-Predictability} \label{sec:formulation}

\subsubsection{$t$-Predictability.}
\add{We consider a task planning problem from a starting state $S\in\mathcal{S}$ with an overall goal $G\in\mathcal{G}$ that can be achieved through a series of actions, called a \textit{plan}, within a finite horizon $T$. 
Let ${\A}$ denote the space of all feasible plans of length (up to) $T$ that achieve the goal.}

\begin{definition}
The $t$-\ACty{}\add{\ $\mathcal{P}_t$} of a \add{feasible plan} $\mathbf{a}=[a_1,a_2,...,a_T]$ that achieves an overall goal $G$ is the probability of an observer correctly inferring $[a_{t+1},...,a_T]$ after observing $[a_1,...,a_t]$, and knowing the overall goal $G$. Specifically, this is given by  ${\mathcal{P}_t(\mathbf{a})=P(a_{t+1},..,a_T|S,G,a_1,..,a_t)}$.
\end{definition}

\begin{definition}
A t-predictable planner generates the plan that maximizes t-predictability out of all those that achieve the overall goal $G$.
That is, a $t$-\AC{} planner generates the action series $\bg^*$ such that $\bg^*=\arg\max_{\bg\in\A}\mathcal{P}_t(\mathbf{a})$.
\end{definition}
This is equivalent, by the general product rule, to:
\begin{equation}
\bg^*=\arg\max_{\bg\in\A} \frac{P(a_1,...,a_T|S,G)}{\sum_{[\tilde{a}_{t+1},...,\tilde{a}_T]}P(a_1,...,a_t,\tilde{a}_{t+1},...,\tilde{a}_T|S,G)}.\label{eq:max_tpred}
\end{equation}

\begin{figure}[t!]
\begin{center}
\includegraphics[width=\textwidth]{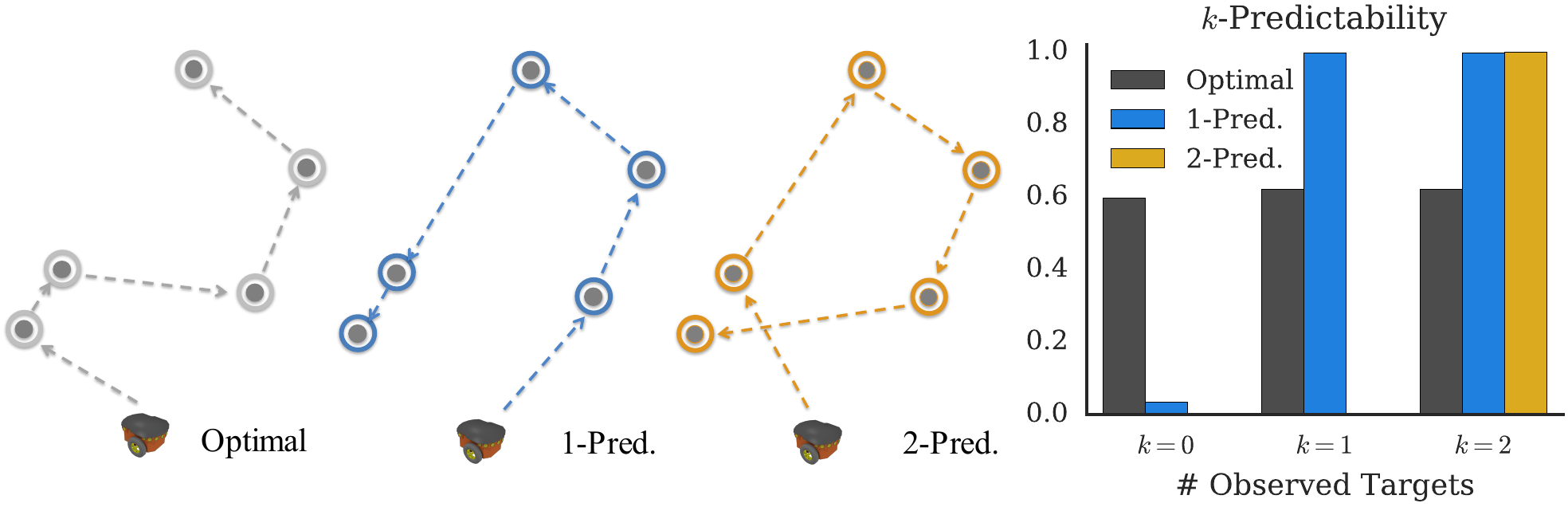}
\caption{\textbf{Theoretical $t$-predictability.} Left: sequences generated by the three planners for a typical task layout.  Right: theoretical $k$-predictability of the same three sequences under different numbers $k$ of observed targets. In all cases, the highest value corresponds to $t=k$.}
\label{fig:ave_theo_pred}
\end{center}
\end{figure}

\subsubsection{Illustrative Example.} 
Fig.~\ref{fig:ave_theo_pred} shows the outcome of optimizing for $t$-\ACty{} in a Traveling Salesman context, with $t=0$, $1$, and $2$ targets, and considers the theoretical $k$-\ACty{} for each plan, with $k$ the number of \emph{actually} observed targets (which may be different from the $t$ assumed by the planner). The $0$-\AC{} plan (gray, left) is the best when the observer sees  no actions, since it is the optimal plan. However, it is no longer the best plan when the observer gets to see the first action: whereas there are multiple low-cost remaining sequences after the first action in the $0$-\AC{} plan, there is only one low-cost remaining sequence after the first action in the $1$-\AC{} plan (blue, center). The first action in the $2$-\AC{} (orange, right) seems irrational, but this plan is optimized for when the observer gets to see the first two actions: indeed, after the first two actions, the remaining plan is maximally clear.

\subsubsection{Relation to Predictability.} $t$-\ACty{} generalizes predictability \cite{Dragan2013}. For $t=0$, the $t$-predictability of a plan simply becomes its predictability, that is, the ease with which the entire sequence of actions can be inferred with knowledge of the overall goal $G$, i.e. 
$\mathcal{P}_0=P(a_{1},..,a_T|S,G)$. 

\subsubsection{Relation to Legibility.} Legibility \cite{Dragan2014} as applied to task planning would maximize the probability of the goal given the beginning of the plan, i.e. 
$ P(G|S,a_1,..,a_t)$. 
In contrast, with $t$-predictability the robot is given a high-level goal describing some state that the world needs to be brought into (for example, clearing all objects from a table), and the observer is \emph{already} aware of this goal. Instead of communicating the goal, the robot conveys the remainder of the plan using the first few elements, maximizing $P(a_{t+1},..,a_T|S,G,a_1,..,a_t)$.

One important implication is that for $t$-predictability, unlike for legibility, \emph{there is no a-priori set entity to be conveyed.} The algorithm searches for \emph{both} a beginning and a remainder of a plan such that, by observing the former, the observer can correctly guess the latter.

Furthermore, legibility and $t$-predictability entail a different kind of information encoding: in legibility, the robot uses a partial trajectory or action sequence to indicate a single goal state, whereas in $t$-predictability the robot uses a partial action sequence to indicate \emph{another} action sequence. Therefore, one entails a mapping from a large space to a small set of possibilities (the finite candidate goal states), whereas the other entails a mapping between spaces of equivalent size.

\add{The distinction between task-level legibility and $t$-\ACty{} is crucially important, particularly in collaborative settings. 
If you are cooking dinner with your household robot, it is important for the robot to act legibly so you can infer \emph{what} goal it has when it grabs a knife (e.g., to slice vegetables).
But, it is equally important for the robot to act in a $t$-\AC{} manner so that you can predict \emph{how} it will accomplish that goal (e.g., the order in which it will cut the vegetables).
}

\subsubsection{Relation to Explicability.} 
Explicability \cite{zhang2016explicability} has been recently introduced to measure whether the observer could assign labels to a plan. In this context, explicability would measure the existence of any remainder of a plan that achieves the goal, as opposed to optimizing the probability that the observer will infer the robot's plan.

\subsubsection{Boltzmann Noisy Rationality.}
Computing $t$-\ACty{} entails computing the conditional probability from \eqref{eq:max_tpred}. We model the human as expecting the robot to be noisily optimal, taking approximately the optimal sequence of actions to achieve $G$. Boltzmann probabilistic models of such noisy optimality (also known as the Luce-Shepard choice rule in cognitive science) have been used in the context of \emph{goal} inference through inverse action planning~\cite{Baker2009}. We adopt an analogous approach for modeling the inference of \emph{task plans}. 

We define optimality via some cost function $c:\A\times \mathcal{S}\times\mathcal{G}\rightarrow \mathbb{R}^+$, mapping each feasible plan, from a starting state and for a particular goal, to a scalar cost. In our experiment, for instance, we use path length (travel distance) for $c$. 
Applying a Bolzmann policy \cite{Baker2009} based on $c$, we get:
\begin{equation}\label{eq:boltzmann}
P(\bg|S,G) = \frac{e^{-\beta c(\bg,S,G)}}{\sum_{\mathbf{\tilde a}\in{\A}}e^{-\beta c(\mathbf{\tilde a},S,G)}}.
\end{equation}
Here $\beta > 0$ is termed the \emph{rationality coefficient}. As $\beta\to\infty$ the probability distribution converges to one for the optimal sequence and zero elsewhere; that is, the human models the agent as rational. As $\beta\to 0$, the probability distribution becomes uniform over all possible sequences $\bg$ and the human models the agent as indifferent.

\subsubsection{$t$-Predictability Optimization.}
We make the assumption that cost is linearly separable, i.e. $c(\bg,S,G)=\sum c(a_t,S^t_\bg,G)$. 
Incorporating \eqref{eq:boltzmann}, \eqref{eq:max_tpred} becomes:
\begin{equation}\label{eq:t_pred}
\mathbf{a^*}=\arg\max_{\bg\in{\A}} \frac{\exp\big(-\beta c(\bg_{t+1:T},S^t_\bg,G)\big)}{\displaystyle\sum_{\tilde{\bg}_{t+1:T}\in{\A}^t_{\bg}}\exp\big(-\beta c(\tilde{\bg}_{t+1:T},S^t_\bg,G)\big)},
\end{equation}
with $S^t_\bg$ denoting the state reached by executing the first $t$ steps of plan $\mathbf{a}$, and ${\A}^t_{\bg}$ denoting the set of all feasible plans that achieve $G$ from state $S^t_\bg$ in $T-t$ steps or less.

\begin{figure}[t!]
\begin{center}
\includegraphics[width=0.66\textwidth,trim=0mm 6mm 0mm 4mm, clip=true]{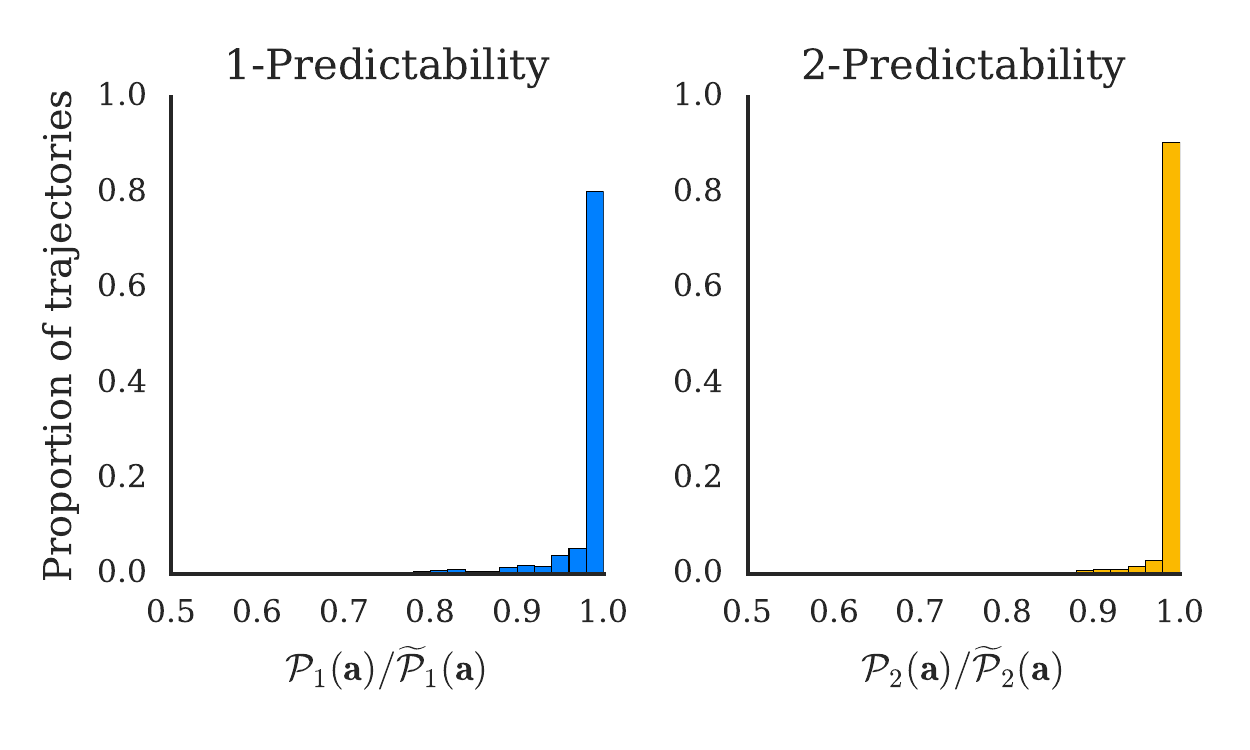}
\caption{\textbf{Approximate $t$-\ACty{}.} Each subplot shows a histogram of ratios between exact $t$-\ACty{} ($\mathcal{P}_t$) and approximate $t$-predictability ($\widetilde{\mathcal{P}}_t$) of all possible sequences across 270 unique layouts. The subplots show histograms of ratios for 1- and 2-\ACty{}, respectively. In the majority of cases, the exact and approximate $t$-predictabilities are nearly identical.}
\label{fig:approx_t_pred}
\end{center}
\end{figure}

\subsubsection{Approximate Algorithm for Large-Scale Optimization.}
\add{
The challenge with the optimization in \eqref{eq:t_pred} is the denominator: it requires summing over all possible plan remainders. Motivated by the fact that plans with higher costs contribute exponentially less to the sum, we propose to approximate the denominator by only summing over the lowest-cost $l$ plan remainders.}

\add{
Many tasks have the structure of Traveling Salesman Problems (TSP), where there are a number of subgoals whose order is not constrained but influences total cost.
}
\add{
Van der Poort et al. \cite{van1999solving} showed how to efficiently compute the $l$ best solutions to the standard (cyclic tour) TSP using a branch-and-bound algorithm.}
\add{The key mechanism is successively dividing the set of feasible plans into smaller subsets for which a lower bound on the cost can be computed by some heuristic.
When a subset of solutions has a lower bound higher than the smallest $l$ costs evaluated so far, it is discarded entirely, while the remaining subsets continue to be broken up. 
The process continues until only $l$ feasible plans remain.
This method is guaranteed to produce the $l$ solutions with the least cost and
can significantly reduce time complexity over exhaustive enumeration. 
In particular, it has been shown that for the standard TSP, computation is in $O(l(T-t)^32^{T-t})$ \cite{van1999solving}%
\add{, while exhaustive enumeration requires computation in $O((T-t)!)$.}
While heuristics are domain-specific,
we expect that this method can be widely applicable to robot task planning problems.}
\add{Further, we expect $T-t$ to be a small number in realistic applications, limited by people's ability to reason about long sequences of actions.}

\add{To empirically evaluate the consequences of this approximation of $t$-\ACty{}, we computed the exact and approximate (using $l=2$) $t$-\ACty{} for all possible plans in 270 randomly generated unique scenes (Fig.~\ref{fig:approx_t_pred}).
If we choose the maximally $t$-\AC{} sequences for each scene using both the exact and approximate calculations of $t$-\ACty{}, we find that these sequences agree in 242 (out of 270) scenes for 1-\ACty{} and in 263 for 2-\ACty{}\footnote{For 0-\ACty{}, the denominator is the same in all plans, so all 270 scenes agree trivially.}.
For the sequences that disagree, the exact $t$-\ACty{} of the sequence chosen using the approximate method is 89.5\% of the optimal $t$-\ACty{} in the worst case, and 99\% of the optimal $t$-\ACty{} on average. This shows that the proposed approximation is highly effective at producing $t$-\AC{} plans.}

%% file: online-experiment.tex

\section{Online Experiment}\label{sec:experiment}

\add{We} set up an experiment to test that our $t$-\AC{} planner is in fact $t$-\AC.
We designed a web-based virtual human-robot collaboration experiment \add{in a TSP setting,} where the human had to predict the behavior of three robot avatars using different planners. Participants watched the robots move to a number of targets (either zero, one, or two) and had to predict the sequence of remaining targets the robot would complete.

\subsection{Independent Variables}

We manipulated two variables: the $t$-\AC{} planner (for $t\in\{0,\ 1,\ 2\}$) and the number of observed targets $k$ (for $k\in\{0,\ 1,\ 2\}$).

\subsubsection{Planner.} We used three different planners which differed in their optimization criterion: the number $t$ of targets assumed known to the observer. Each participant interacted with three robot avatars, each using one of the following three planners:

\noindent\emph{Optimal ($0$-\AC):} This robot chooses the shortest path from the initial location visiting all target locations once; that is, the ``traditional'' solution to the open TSP. This robot solves \eqref{eq:t_pred} for $t=0$.

\noindent\emph{1-\AC:} This robot solves \eqref{eq:t_pred} for $t=1$; the sequence might make an inefficient choice for the first target in order to make the sequence of remaining targets clear.

\noindent\emph{2-\AC:} This robot solves \eqref{eq:t_pred} for $t=2$; the sequence might make an inefficient choice for the first \emph{two} targets in order to make the sequence of remaining targets clear.

\subsubsection{Number of observed targets.} Each subject was shown the first $k\in\{0,1,2\}$ targets of the robot's chosen sequence in each trial and was asked to predict the remainder of the sequence. This variable was manipulated between participants; thus, a given participant always saw the same number $k$ of initial targets on all trials.

\subsection{Procedure}

The experiment was divided into two phases: a training phase to familiarize participants with TSPs and how to solve them, and an experimental phase. We additionally asked participants to fill out a survey at the end of the experiment.

In the training phase, subjects controlled a human avatar. They were instructed to click on targets in the order that they believed would result in the quickest path for the human avatar to visit all of them. The human avatar moved in a straight line after each click and ``captured'' the selected target, which was then removed from the display. 

For the second phase of the experiment, participants saw a robot avatar move to either $k=0$, $k=1$, or $k=2$ targets. After moving to these targets, the robot paused so that participants could predict the remaining sequence of targets by clicking on the targets in the order in which they believed the robot would complete them. Afterwards, participants were presented with an animation showing the robot moving to the rest of the targets in the sequence determined by the corresponding planner.

\subsubsection{Stimuli.} Each target layout displayed a square domain with five or six targets. There were a total of 60 trials, consisting of four repetitions of 15 unique target layouts in random order: one for the training phase, in addition to the three experimental conditions. The trials were grouped so that each participant observed the same robot for three trials in a row before switching to a different robot. In the training trials, the avatar was a gender-neutral cartoon of a person on a scooter, and the robot avatars were images of the same robot in different poses and colors (either red, blue, or yellow).

\subsubsection{Layout Generation.} The layouts for the 15 trials were based from an initial database of 270 randomly generated layouts.
This number was reduced down to 176 in which the chosen sequence was different between all three planners so that the stimuli were distinguishable.
We also discarded some scenarios in which the robot's trajectory approached a target without capturing it, to avoid confounds.
Out of these valid layouts, we chose the ones with the highest theoretical gain in 1-predictability to 2-predictability, to avoid scenarios where the information gain was marginal.

\subsubsection{Attention Checks.} After reading the instructions, participants were given an attention check in the form of two questions asking them the color of the targets and the color of the robot that they \textit{would not} be evaluating.

\subsubsection{Controlling for Confounds.} We controlled for confounds by counterbalancing the colors of the robots for each planner; by using a human avatar in the practice trials; by randomizing the trial order; and by including attention checks.

\subsection{Dependent Measures}

\subsubsection{Objective measures.} We recorded the proportion of correct predictions of the robot's sequence of targets out of all 15 trials for each planner, resulting in a measure of \textit{error rate}. We additionally computed the \textit{Levenshtein distance} between predicted and actual sequences of targets. This is a more precise measure of how similar participants' predictions were to the actual sequences produced by the planner.

\subsubsection{Subjective measures.} After every ninth trial of the experiment, we asked participants to indicate which robot they preferred working with. At the end the experiment, each participant was also asked to complete a questionnaire to evaluate their perceived performance of three robots. An informal analysis of this questionnaire suggested similar results as those obtained from our other measures (see Section~\ref{sec:online-results}). Thus, because of space constraints, we have omitted specifics of the survey in this paper.

\subsection{Hypotheses}

\noindent\textbf{H1 - Comparison with Optimal.} \emph{When showing 1 target, the $1$-\AC{} robot will result in lower error than the optimal baseline. When showing 2 targets, the $2$-\AC{} robot will result in lower error than the optimal baseline.}

\noindent\textbf{H2 - Generalization.} \textit{The error rate will be lowest when $t=k$: the number of targets shown, $k$, equals the number of targets assumed by the $t$-\AC{} planner, $t$.}

\noindent\textbf{H3 - Preference.} \textit{The perceived performance of the robots will be highest when $t=k$.}

\subsection{Participants}

We recruited a total of 242 participants from Amazon's Mechanical Turk using the psiTurk experimental framework \cite{Gureckis15}. We excluded 42 participants from analysis for failing the attention checks, leaving a net total of $N=200$ participants.
All participants were treated in accordance with local IRB standards and were paid \$1.80 USD for an average of 22 minutes of work, plus an average performance-based bonus of \$0.47.

\input{online-results.tex}

%% file: online-results.tex

\subsection{Results}
\label{sec:online-results}

\begin{figure}[t!]
\begin{center}
\includegraphics[width=\textwidth,trim=0mm 5mm 0mm 5mm, clip=true]{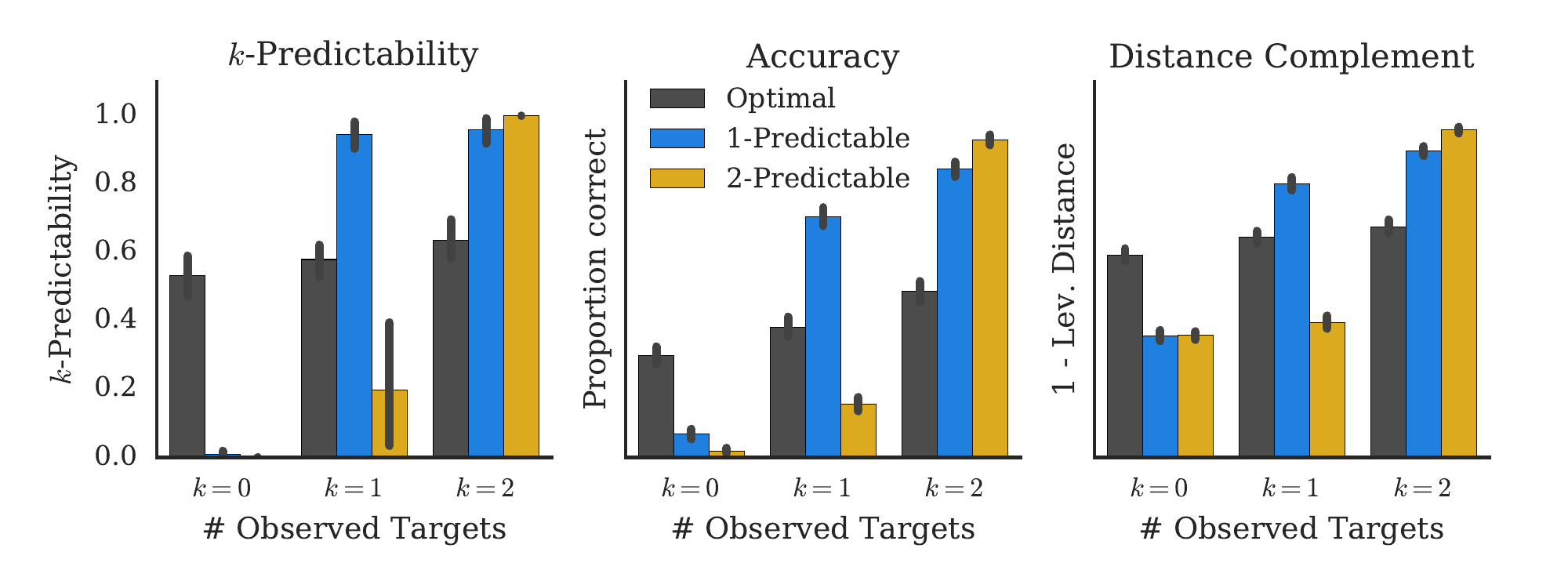}
\caption{\textbf{Predictability, error rate and edit distance.} Left: theoretical $k$-predictability of sequences generated by different $t$-\AC{} planners under different numbers $k$ of observed targets, averaged over all task layouts used in the online experiment. In all cases, the highest value corresponds to $t=k$. Center: empirical proportion of correct predictions with different $t$-\AC{} planners for different numbers $k$ of observed targets. Right: complement of the average empirical Levenshtein distance between predicted and correct sequences. The lowest experimental error rates under both metrics occur when $t=k$.}
\label{fig:accuracy}
\end{center}
\end{figure} 

\subsubsection{Model validity.} \label{sec:model-validity} We first looked at the validity of our model of $t$-\ACty{} with respect to people's performance in the experiment. We computed the theoretical $k$-\ACty{} (probability of correctly predicting the robot's sequence from the $k$ targets the user saw) for each task layout under each planner and number of targets the users observed. We also computed people's actual prediction accuracy on each of these layouts under each condition, averaged across participants. 

We computed the Pearson correlation between $k$-\ACty{} and participant accuracy, finding a correlation of \PredictabilityVsAccuracy{}; the confidence interval around the median was computed using 10,000 bootstrap samples (with replacement). \textbf{\emph{This high correlation suggests that our model of how people predict action sequences of other agents is a good predictor of their actual behavior.}}

\subsubsection{Accuracy.} To determine how similar people's predictions of the robots' sequences were to the actual sequences, we used two objective measures of accuracy: first, overall error rate (whether they predicted the correct sequence or not), as well as the Levenshtein distance between the predicted and correct sequences (Fig.~\ref{fig:accuracy}). 

As the two measures have qualitatively similar patterns of results, and the Levenshtein distance is a more fine-grained measure of accuracy, we performed quantitative analysis only on the Levenshtein distance. We constructed a linear mixed-effects model with the number of observed targets $k$ ($k$ from 0 to 2) and the planner for $t$-\ACty{} ($t$ from 0 to 2) as fixed effects, and trial layout as random effects.

This model revealed significant main effects of the number of observed targets (\AccuracyConditionEffect{}) and planner (\AccuracyRobotEffect{}) as well as an interaction between the two (\AccuracyInteraction{}). We ran post-hoc comparisons using the multivariate $t$ adjustment. Comparing the planners across the same number of targets, we found that
in the 0-targets condition the \ACzero{} (or 0-predictable) robot was better than the other two robots; in the 1-target condition, the \ACone{} robot was better than the other two; in the 2-target prediction, the \ACtwo{} robot was better than the optimal and \ACone{} robots.
All differences were significant with $p<0.001$ except the difference between the \ACtwo{} robot and the \ACone{} robot in the 2-target condition (\AccuracyConditionTwoLegibleOnevsLegibleTwo{}). Comparing the performance of a planner across number of targets, we found significant differences in all contrasts, with one exception: the accuracy when using the optimal planner was not significantly different when seeing 1 target vs 2 targets (\AccuracyOptimalCondOnevsCondTwo{}).
Overall, these results support our hypotheses \textbf{H1} and \textbf{H2}, that \textbf{\emph{accuracy is highest when $t$ used in the planner equals $k$, the number of observed targets.}}

\subsubsection{Preferences over time.} 
\begin{figure}[t!]
\begin{center}
\includegraphics[width=0.9\textwidth,trim=0mm 4.5mm 0mm 5mm, clip=true]{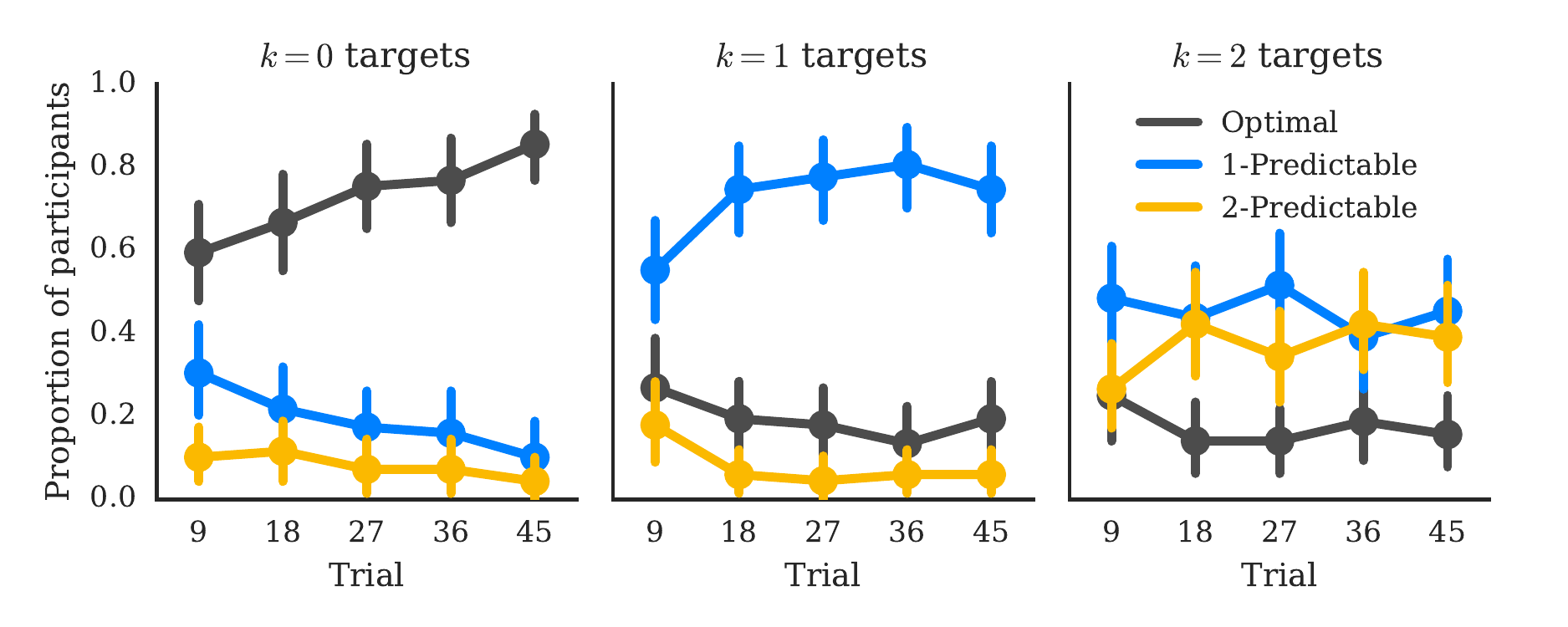}
\caption{\textbf{Preferences over time.} Participants prefer the $0$-\AC{} (optimal) robot for $k=0$ and the $1$-\AC{} robot for $k=1$, as well as $k=2$ (despite performing  better with the $2$-\AC{} robot, subjects often report being confused or frustrated by its first 2 actions.)}
\label{fig:preferences-over-time}
\end{center}
\end{figure}

Fig.~\ref{fig:preferences-over-time} shows the proportion of participants choosing each robot planner at each trial. We constructed a logistic mixed-effects model for binary preferences (where 1 meant the robot was chosen) with planner, number of observed targets, and trial as fixed effects and participants as random effects. The planner and number of observed targets were categorical variables, while trial was a numeric variable.

Using Wald's tests, we found a significant main effect of the planner (\PreferenceRobot{}) and trial (\PreferenceTrial{}). We detected only a marginal effect of number of targets (\PreferenceCondition{}). However, there was a significant interaction between planner and number of targets (\PreferenceRobotCondition{}). We also found interactions between planner and trial (\PreferenceRobotTrial{}) and between number of targets and trial (\PreferenceConditionTrial{}), as well as a three-way interaction (\PreferenceRobotConditionTrial{}).
Post-hoc comparisons with the multivariate $t$ adjustment for $p$-values indicated that for the 0-targets condition, the \ACzero{} robot was preferred over the \ACone{} robot (\PrefsConditionZeroLegibleOnevsOptimal{}) and the \ACtwo{} robot (\PrefsConditionZeroLegibleTwovsOptimal{}). For the 1-target condition, the \ACone{} robot was preferred over the \ACzero{} robot (\PrefsConditionOneLegibleOnevsOptimal{}) and the \ACtwo{} robot (\PrefsConditionOneLegibleOnevsLegibleTwo{}). In the two-task condition, we did not detect a difference between the two \ACone{} and \ACtwo{} robots (\PrefsConditionTwoLegibleOnevsLegibleTwo{}), though both were preferred over the \ACzero{} robot (\PrefsConditionTwoLegibleOnevsOptimal{} for the \ACone{} robot and \PrefsConditionTwoLegibleTwovsOptimal{} for the \ACtwo{} robot).

Overall, these results are in line with our hypothesis \textbf{H3} that \textbf{\emph{the perceived performance is highest when $t$ used in the planner equals $k$, the number of observed targets.}} This is the case for $k=0$ and $k=1$, but not $k=2$: even though users tended to perform better with the $2$-\AC{} robot, its suboptimal actions in the beginning seemed to confuse and frustrate users (see Qualtitative feedback results for details).

\subsubsection{Final rankings.} 
\begin{figure}[t!]
\begin{center}
\includegraphics[width=0.9\textwidth,trim=0mm 5mm 0mm 4mm, clip=true]{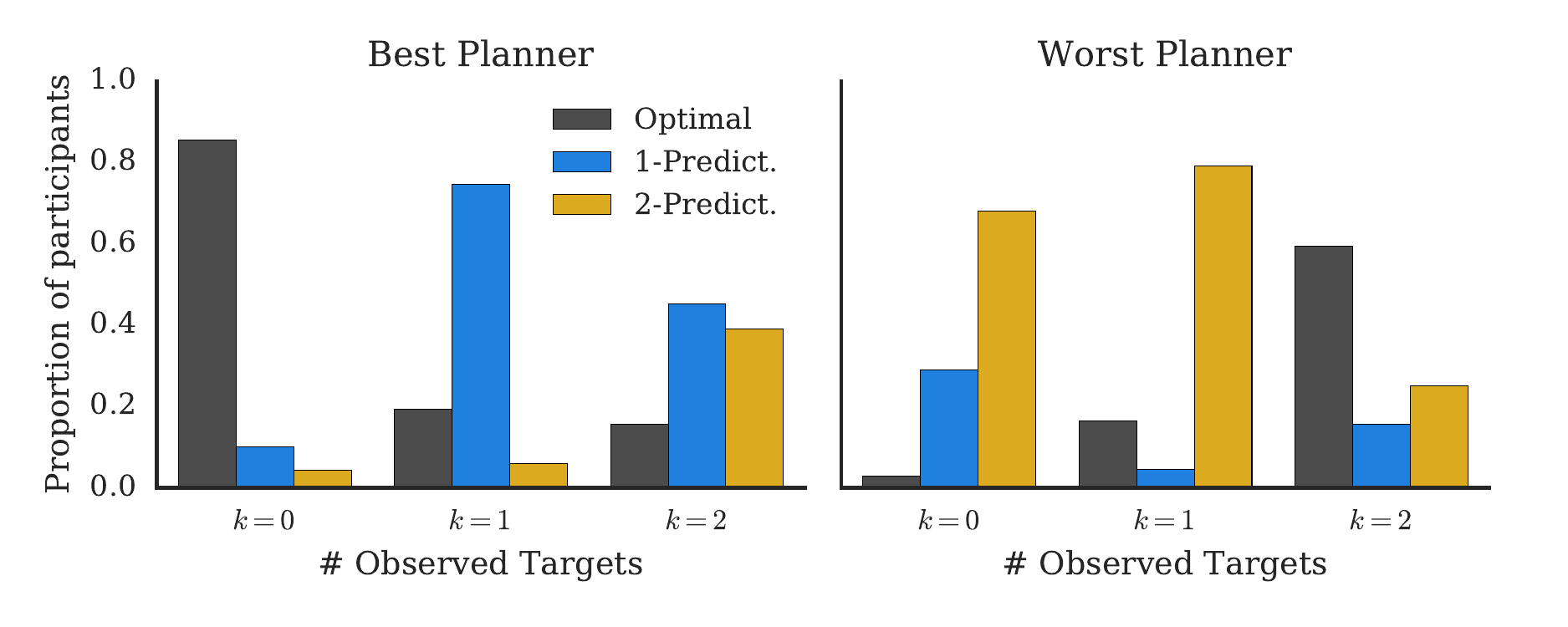}
\caption{\textbf{Final rankings.} Participants ranked the planners differently depending on how many targets were observed. For $k=0$, people preferred the optimal planner; for $k=1$ and $k=2$, they preferred the 1-\AC{} planner.}
\label{fig:rankings}
\end{center}
\end{figure}

The final rankings of ``best robot'' and ``worst robot'' are shown in Fig.~\ref{fig:rankings}.
For each participant, we assigned each robot a score based on their final rankings. The best robot received a score of 1; the worst robot received a score of 2; and the remaining robot received a score of 1.5. We constructed a logistic mixed-effects model for these scores, with planner and number of observed targets as fixed effects, and participants as random effects; we then used Wald’s tests to check for effects. 

We found significant main effects of planner (\RankingsRobot{}) and number of targets (\RankingsCondition{}), as well as an interaction between them (\RankingsRobotCondition{}). We again performed post-hoc comparisons using the multivariate $t$ adjustment. These comparisons indicated that in the 1-target condition, people preferred the \ACzero{} robot over the \ACone{} robot (\RankingsConditionZeroLegibleOnevsOptimal{}) and the \ACtwo{} robot (\RankingsConditionZeroLegibleTwovsOptimal{}). In the 1-target condition, there was a preference for the \ACone{} robot over the \ACzero{} robot, however this difference was not significant (\RankingsConditionOneLegibleOnevsOptimal{}). The \ACone{} robot was preferred to the \ACtwo{} robot (\RankingsConditionOneLegibleOnevsLegibleTwo{}). In the 2-target condition, both the \ACone{} and \ACtwo{} robots were preferred over the \ACzero{} robot (\RankingsConditionTwoLegibleOnevsOptimal{} for the \ACone{} robot, and \RankingsConditionTwoLegibleTwovsOptimal{} for the \ACtwo{} robot), though we did not detect a difference between the the \ACone{} and \ACtwo{} robots themselves (\RankingsConditionTwoLegibleOnevsLegibleTwo{}). Overall, these rankings are in line with the preferences over time.

\subsubsection{Qualitative feedback.} At the end of the experiment, we asked participants to briefly comment on each robot. For $k=0$, responses typically favored the optimal robot, often described as ``efficient" and ``logical", although they also showed some reservations:
\emph{``close to what I would do but just a little bit of weird choices tossed in''}.
Conversely, for $k>0$, the optimal robot was likened to ``a dysfunctional computer'',
and described as ``ineffective'' or ``very robotic": \emph{``I feel like maybe I'm a dumb human and the [optimal] robot might be the most efficient, because I have no idea. It frustrated me.''}

The \ACtwo{} robot had mixed reviews for $k=2$: for some it was ``easy to predict'',
others found it
``misleading" or noted its ``weird starting points''. For $k<2$, it was reported as ``useless'', ``all over the place'', and \emph{``terribly unintuitive with an abysmal sense of planning"}; one participant wrote it \emph{``almost seemed like it was trying to trip me up on purpose''}
and another one declared
\emph{``I want to beat this robot against a wall.''}

The \ACone{} robot seemed to receive the best evaluations overall: though for $k=0$ many users found it ``random'', ``frustrating'' and ``confusing'', for $k>0$ it almost invariably had positive reviews (``sensible", ``reasonable'', ``dependable", ``smart'', ``on top of it"), being likened to ``a logical and rational human'' and even eliciting positive emotions: \emph{``You're my boy, Blue!''}, \emph{``I like the little guy, he thinks like me''}, or \emph{``It was my favorite. I started thinking `Don't betray me, Yellow!' as it completed its sequence.''}

\subsubsection{Summary.} Our $t$-\ACty{} planner worked as expected, with the $t$-\AC{} robots leading to the highest user prediction accuracy given the first $t$ targets. However, focusing on just $2$-\ACty{} at the expense of $0$-\ACty{} frustrated our users. Overall, we believe $t$-\ACty{} will be important in a task for all $t$s, and hypothesize that optimizing for a weighted combination will perform best in practice. 

We note that $\beta$ is problem-specific and can be expected to decay as the difficulty of the task increases; in each setting, it can be estimated from participant data. 
Although $\beta$ was chosen ahead of time in our experiment to be $\beta=1$, our results are  validated by the \PredictabilityVsAccuracyShort{} correlation between expected and observed human error rates.

The optimal choice of $t$ is also a subject for further investigation and is likely context-specific. Depending on the particular task, there should be a different trade-off between predictability of later actions and that of earlier actions.

%% file: user-study.tex

\section{User Study}
Having tested our ability to produce $t$-\AC{} sequences, we next ran an in person study to test their implications. Participants used a smartphone to operate a remote-controlled Sphero BB-8 robot, and had to predict and adapt to the actions of an autonomous Pioneer P3DX robot in a collaboration scenario (Fig.~\ref{fig:front}).

\subsection{Independent Variables}

We manipulated one single variable, \emph{planner}, as a within-subjects factor. Having confirmed the expected effects of the different planners in the previous experiment, and given the good overall performance of the 1-predictable planner across different conditions, we decided to omit the 2-predictable agent and focus on testing the implications of 1-predictable with respect to optimal in a more immersive collaborative context.

\subsection{Procedure}

At the beginning of the experiment, participants were told that they and their two robot friends were on a secret mission to deactivate an artifact. In each of 4 trials, the autonomous P3DX navigated to the 5 power sources and deactivated them in sequence; however, security sensors activated at each power source after 3 or more had been powered down. The subject's mission was to use BB-8 to jam the sensors at the third, fourth and fifth power sources before the P3DX arrived at them, by steering BB-8 into the corresponding sensor for a short period of time. 

After an initial practice phase in which participants had a chance to familiarize themselves with the objective and rules of the task, as well as with the BB-8 teleoperation interface, there were two blocks of 4 trials whose order was counterbalanced across participants. In each block, the subject collaborated with the P3DX under a different task planner which we referred to as different robot ``personalities''.

\subsubsection{Stimuli.} Each of the 5 power sources (targets) in each trial was projected onto the floor as a yellow circle, using an overhead projector (Fig~\ref{fig:front}). Each circle was initially surrounded by a projected blue ring representing a dormant sensor. When the P3DX reached a target, both the circle and the ring were eliminated, except When the P3DX reached the third target, in which case the blue circles turned red symbolizing their switch into active state. Whenever BB-8 entered a ring, the ring turned green for 2 seconds and then disappeared, indicating successful jamming. If the P3DX moved over a red ring, a large red rectangle was projected, symbolizing capture and the trial ended in failure. Conversely, if the P3DX completed all 5 targets without entering a red ring, a green rectangle indicated successful completion of the trial.

\subsubsection{Layout Generation.} The 4 layouts used were taken from the larger pool of 15 layouts in the online experiment. There was a balance between layouts where online participants had been more accurate with the optimal planner, more accurate with the 1-predictable planner, or similarly accurate.

\subsubsection{Controlling for Confounds.} We controlled for confounds by counterbalancing the order of the planners; by using a practice layout; and by randomizing the trial order.

\subsection{Dependent Measures}

\subsubsection{Objective measures.} We recorded the number of successful trials for each subject and robot planner, as well as the number of trials where participants jammed targets in the correct sequence.

\subsubsection{Subjective measures.} After every block of the experiment, each participant was also asked to complete a questionnaire (adapted from \cite{Dragan2015}) to evaluate their perceived performance of the P3DX robot.
At the end of the experiment, we asked participants to indicate which robot (planner) they preferred working with.

\subsection{Hypotheses}

\noindent\textbf{H4 - Comparison with Optimal.} \emph{The $1$-\AC{} robot will result in more successful trials than the optimal baseline.}

\noindent\textbf{H5 - Preference.} \textit{Users will prefer working with the $1$-\AC{} robot.}

\subsection{Participants}

We recruited 14 participants from the UC Berkeley community, who were treated in accordance with local IRB standards and paid \$10 USD. The study took about 30~min.

\input{user-study-results.tex}

%% file: user-study-results.tex

\subsection{Results}

\subsubsection{Successful completions.} We first looked at how often participants were able to complete the task with each robot. We constructed a logistic mixed-effects model for completion success with planner type as a fixed effect and participant and task layout as random effects. We found a significant effect of planner type (\CompletionRobot{}), with the \ACone{} robot yielding more successful completions than the \ACzero{} robot (\CompletionLegibleOnevsOptimal{}). This supports \textbf{H4}.

\subsubsection{Prediction accuracy.} We also looked at how accurate participants were at predicting the robots' sequence of tasks, based on the order in which participants jammed tasks.
We constructed a logistic mixed-effects model for prediction accuracy with planner type as a fixed effect and participant and task layout as random effects. We found a significant effect of planner type (\UserStudyAccuracyRobot{}), with the \ACone{} robot being more predictable than the \ACzero{} robot (\UserStudyAccuracyLegibleOnevsOptimal{}).

\subsubsection{Robot preferences.} We asked participants to pick the robot they preferred to collaborate with. We found that 86\% ($N=12$) of participants preferred the \AC{} robot, while the rest ($N=2$) preferred the \ACzero{} robot. This result is significantly different from chance (\UserStudyPreferencesRobot{}). This supports \textbf{H5}.

\begin{figure}[t!]
\begin{center}
\includegraphics[width=0.72\textwidth,trim=0mm 4mm 0mm 4mm, clip=true]{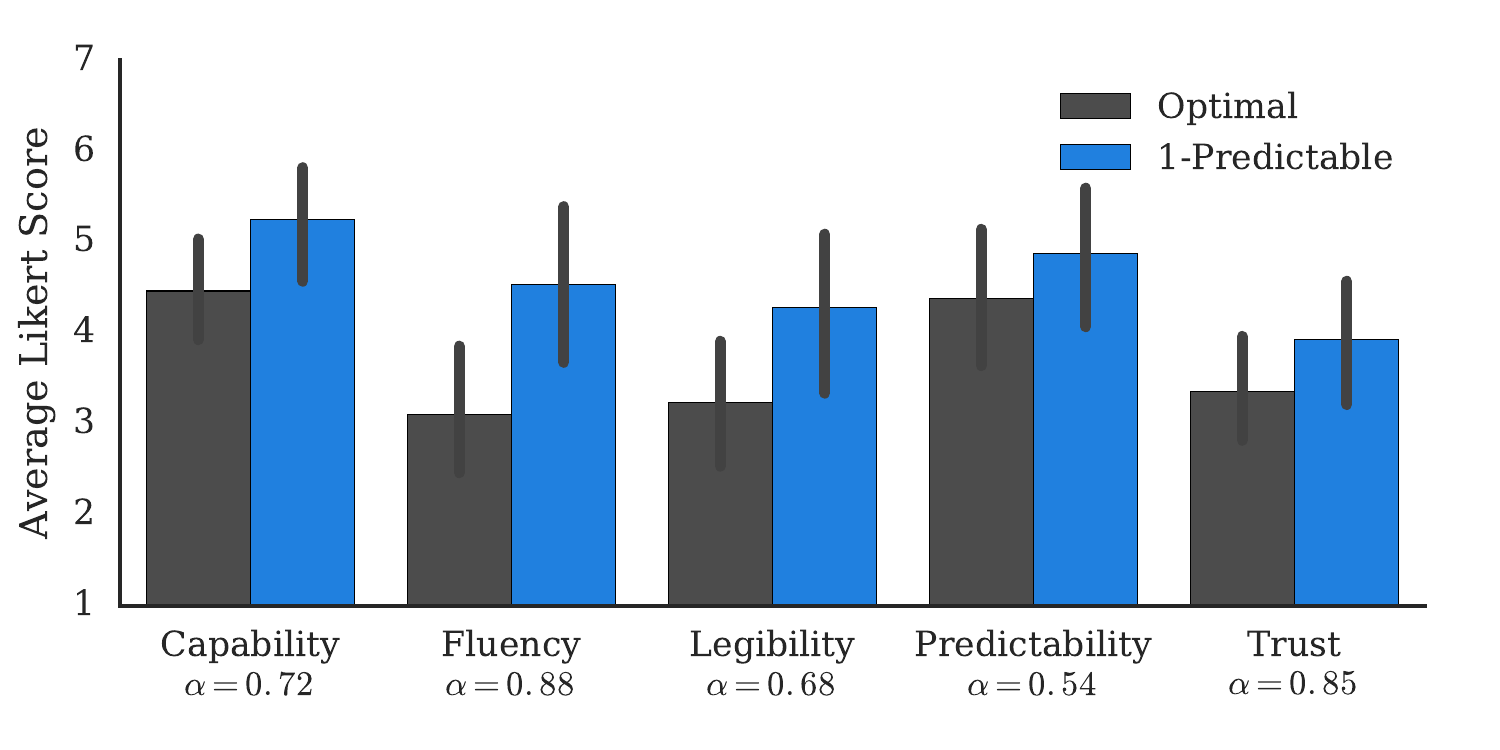}
\caption{\textbf{Perceptions of the collaboration.} Over all measures, participants ranked the 1-\AC{} planner as being preferable to the optimal planner.}
\label{fig:user-study-survey}
\end{center}
\end{figure}

\subsubsection{Perceptions of the collaboration.}
We analyzed participants' perceptions of the robots' behavior (Fig.~\ref{fig:user-study-survey}) by averaging each participant's responses to the individual questions for each robot and measure, resulting in a single score per participant, per measure, per robot. We constructed a linear mixed-effects model for the survey responses with planner and measure type as fixed effects, and with participants as random effects. We found a main effect of planner (\UserStudySurveyRobot{}) and measure (\UserStudySurveyMeasure{}). Post-hoc comparisons using the multivariate $t$ method for $p$-value adjustment indicated that participants preferred the \AC{} robot over the \ACzero{} robot (\UserStudySurveyLegibleOnevsOptimal{}) by an average of $0.87\pm 0.21\ \mathrm{SE}$ points on the Likert scale.

%% file: discussion.tex

\section{Discussion}

In what remains, we summarize our work and discuss future directions, including the application of $t$-\ACty{} beyond task planning.

\subsubsection{Summary.} We enable robots to generate $t$-\AC{} plans, for which a human observing the first $t$ actions can confidently infer the rest. We tested the ability to make plans $t$-predictable in a large-scale online experiment, in which subjects' predictions of the robot's action sequence significantly improved. In an in-person study, we found that $t$-\ACty{} can lead to significant objective and perceived improvements in human-robot collaboration compared to traditional optimal planning.

\subsubsection{$t$-\ACty{} for Motion.}
Even though $t$-\ACty{} is motivated by a task planning need, it does have an equivalent in motion planning: \add{find an initial trajectory $\xi_{0:t}$, such that the remainder $\xi_{t:T}$ can be inferred by a human observer with knowledge of both the start state $\xi_0=S$ and the goal state $\xi_T=G$.}
\add{Modeling this conditional probability with a Boltzmann model yields
\begin{equation*}
 \xi^*=\arg\max_{\xi\in\Xi} P(\xi_{t:T}|G,\xi_{0:t})=\arg\max_{\xi\in\Xi}\frac{e^{-\beta c(\xi_{t:T})}}{\int e^{-\beta c(\hat{\xi}_{t:T})}d\hat{\xi}_{t:T}},
\end{equation*}
where $\Xi$ is the set of feasible trajectories from $S$ to $G$. Using a second order expansion of the cost $c(\hat{\xi}_{t:T})$ about the optimal remaining trajectory, we get:
\begin{align}
\label{eq:tpred_hessian}
\max_{\xi}P(\xi_{t:T}|G,\xi_{0:t})
&\approx \max_{\xi_{0:t}}\frac{e^{-\beta c(\xi^*_{t:T})}}{e^{-\beta c(\xi^*_{t:T})}\int e^{-\frac{1}{2}(\hat{\xi}_{t:T}-\xi^*_{t:T})^T H(\xi^*_{t:T})(\hat{\xi}_{t:T}-\xi^*_{t:T})}d\hat{\xi}_{t:T}}\nonumber\\
& \equiv\max_{\xi_{t}}\det(H(\xi^*_{t:T})),
\end{align}}%
where $H$ denotes the Hessian.
This implies that generating a $t$-predictable trajectory means finding a configuration for time $t$ such that the optimal trajectory from that configuration to the goal is in a \emph{steep}, high-curvature minimum: other trajectories from $\xi_t$ to the goal would be significantly higher cost. For instance, if the robot had the option between two passages, it would choose the more \emph{narrow} passage because that enables the observer to more accurately predict the remainder of its trajectory.

\subsubsection{Limitations and Future Work.} 
Our work is limited by the focus in our experiments on TSP scenarios (though we emphasize that $t$-\ACty{} as it is formulated in Section~\ref{sec:formulation} is not inherently limited to TSPs).
This work is also limited by the choice of a user study that involved a tele-operated avatar to mediate the human's physical collaboration with the robot.
Applications to other scenarios that involve direct physical collaboration and preconditions would be interesting topics to investigate.
Additionally, while our work showcases the utility of $t$-\ACty, a main remaining challenge is determining what $t$ or combination of $t$s to use for arbitrary tasks.
\add{This decision requires more sophisticated human behavior models, which are the topic of our ongoing work.}